\documentclass[10pt,twocolumn,letterpaper]{article}

\usepackage{cvpr}
\usepackage{times}
\usepackage{epsfig}
\usepackage{graphicx}
\usepackage{amsmath}
\usepackage{amssymb}
\usepackage{nicefrac}       
\usepackage{comment}
\usepackage{booktabs}

\usepackage{soul} 

\newcommand{\x}{\mathbf{x}}
\newcommand{\xtilde}{\mathbf{\tilde{x}}}
\newcommand{\xhat}{\mathbf{\hat{x}}}

\newcommand{\z}{\mathbf{z}}

\newcommand{\Xcal}{\mathcal{X}}
\newcommand{\Ycal}{\mathcal{Y}}
\newcommand{\Lcal}{\mathcal{L}}

\newcommand{\PP}{\mathbb{P}} 

\newcommand{\eq}[1]{(\ref{#1})}
\newcommand{\RR}{\mathbb{R}}

\newcommand{\inner}[2]{\left\langle #1,#2 \right\rangle}
\def\ci{\perp\!\!\!\perp}
\newcommand{\tr}{\mathop{\mathrm{tr}}}
\newcommand*{\mini}{\mathop{\mathrm{minimize}}}

\usepackage{amsthm}
\usepackage{framed}
\newcommand{\BOXAMPLE}{Example}
\newcommand{\BOXDEF}{Definition}

\newtheoremstyle{nospace}
     {0pt}
     {0pt}
     {}
     {}
     {\bfseries}
     {:}
     {.5em}
     {}

\theoremstyle{nospace}
\newtheorem{example}{\BOXAMPLE}
\newtheorem{definition}{\BOXDEF}

\newenvironment{fshaded}{%
\MakeFramed {\FrameRestore}}%
{\endMakeFramed}
\usepackage{xcolor}
\colorlet{shadecolor}{black!10}
\colorlet{framecolor}{black}

\usepackage[pagebackref=true,breaklinks=true,letterpaper=true,colorlinks,bookmarks=false]{hyperref}

\cvprfinalcopy 

\def\cvprPaperID{3708} 

\ifcvprfinal\fi
\begin{document}

\title{Discovering Fair Representations in the Data Domain}

\author{Novi Quadrianto$^\ddag$\thanks{Also with Higher School of Economics, Moscow, Russia} , Viktoriia Sharmanska$^\mathsection$, Oliver Thomas$^\ddag$\\
$\ddag$Predictive Analytics Lab (PAL), University of Sussex, Brighton, United Kingdom\\
$\mathsection$Department of Computing, Imperial College London, United Kingdom\\
%
}

\maketitle
\thispagestyle{empty}

\begin{abstract}
Interpretability and fairness are critical in computer vision and machine learning applications, in particular when dealing with human outcomes, e.g. inviting or not inviting for a job interview based on application materials that may include photographs.
One promising direction to achieve fairness is by learning data representations that remove the semantics of protected characteristics, and are therefore able to mitigate unfair outcomes.
All available models however learn latent embeddings which comes at the cost of being uninterpretable.
We propose to cast this problem as data-to-data translation, i.e. learning a mapping from an input domain to a fair target domain, where a fairness definition is being enforced.  
Here the data domain can be images, or any tabular data representation. 
This task would be straightforward if we had fair target data available, but this is not the case. 
To overcome this, we learn a highly unconstrained mapping by exploiting statistics of residuals -- the difference between input data and its translated version -- and the protected characteristics. 
When applied to the CelebA dataset of face images with gender attribute as the protected characteristic, our model enforces equality of opportunity by adjusting the eyes and lips regions.
Intriguingly, on the same dataset we arrive at similar conclusions when using semantic attribute representations of images for translation.  
On face images of the recent DiF dataset, with the same gender attribute, our method adjusts nose regions.
In the Adult income dataset, also with protected gender attribute, our model achieves equality of opportunity by, among others, obfuscating the wife and husband relationship.
Analyzing those systematic changes will allow us to scrutinize the interplay of fairness criterion, chosen protected characteristics, and prediction performance. 
\end{abstract}

\section{Introduction}
\label{sec:intro}
Machine learning systems are increasingly used by government agencies, businesses, and other organisations to assist in making life-changing decisions such as whether or not to invite a candidate to a job interview, or whether to give someone a loan. 
%
The question is how can we ensure that those systems are \emph{fair}, i.e. they do not discriminate against individuals because of their gender, disability, or other personal (``protected'') characteristics?
For example, in building an automated system to review job applications, a photograph
might be used in addition to other features to make an invite decision. 
By using the photograph as is, a discrimination issue might arise, as photographs with faces could reveal 
 certain protected characteristics, such as gender, race, or age (e.g. \cite{FuHeHou2014, BruBurHan93,brown1993,LevHas15}).
Therefore, any automated system that incorporates photographs into its decision process is at risk of indirectly conditioning on protected characteristics (indirect discrimination).
%
%
Recent advances in learning fair representations suggest adversarial training as the means to hide the protected characteristics from the decision/prediction function \cite{BeuCheZhaChi17,ZhaLemMit18,MadCrePitZem18}. 
All fair representation models, however, learn \emph{latent embeddings}. 
Hence, the produced representations cannot be easily interpreted. They do not have the semantic meaning of the input that photographs, or education and training attainments, provide when we have job application data. 
If we want to encourage public conversations and productive public debates regarding fair machine learning systems \cite{WEF18}, interpretability in how fairness is met is an integral yet overlooked ingredient.

In this paper we focus on representation learning models that can transform inputs to their fair representations and retain the semantics of the input domain in the transformed space.
When we have image data, our method will make a semantic change to the appearance of an image to deliver a certain fairness criterion\footnote{Examples of fairness criteria are equality of true positive rates (TPR), also called equality of opportunity \cite{HarPriSre16,ZafValRodGum17b}, between males and females.}. 
To achieve this, we perform \emph{a data-to-data translation} by learning a mapping from data in a source domain to a target domain.
Mapping from source to target domain is a standard procedure, and many methods are available. 
For example, in the image domain, if we have aligned source/target as training data, we can use the pix2pix method of \cite{IsoZhuZhoEfr17}, which is based on conditional generative adversarial networks (cGANs) \cite{MirOsi14}. 
Zhu et al.'s CycleGAN \cite{ZhuParIsoEfr17} and Choi et al.'s StarGAN \cite{ChoChoKimHaetal18} solve a more challenging setting in which only \emph{un}aligned training examples are available. 
However, we can not simply reuse existing methods for source-to-target mapping because we do \emph{not have data in the target domain} (e.g. fair images are not available; images by themselves can not be fair or unfair, it is only when they are coupled with a particular task that the concern of fairness arises). 

To illustrate the difficulty, consider our earlier example of an automated job review system that uses photographs as part of an input. 
For achieving fairness, it is tempting to simply use GAN-driven methods to \emph{translate female face photos to male}. 
We would require training data of female faces (source domain) and male faces (target domain), and only unaligned training data would be needed. 
This solution is however fundamentally flawed; who gets to decide that we should translate in this direction?
Is it fairer if we translate male faces to female instead?
An ethically grounded approach would be to translate both male and female face photos (source domain) to appropriate middle ground face photos (target domain).
This challenge is actually multi-dimensional, it contains at least \emph{two sub-problems}: a) how to have a general approach that can handle image data as well as tabular data (e.g. work experience, education, or even semantic attribute representations of photographs), and b) how to find a middle-ground with a multi-value (e.g. race) or continuous value (e.g. age) protected characteristic or even multiple characteristics (e.g. race and age).

%
%
We propose a solution to the multi-dimensional challenge described above by exploiting statistical (in)dependence between translated images and protected characteristics. 
We use the Hilbert-Schmidt norm of the cross-covariance operator between reproducing kernel Hilbert spaces of image features and protected characteristics (Hilbert-Schmidt independence criterion \cite{GreBouSmoSch05}) as an empirical estimate of statistical independence.
This flexible measure of independence allows us to take into account higher order independence, and handle a multi-/continuous value and multiple protected characteristics.

\noindent\textbf{Related work}
We focus on expanding the related topic of learning fair, \emph{albeit uninterpretable}, representations.
The aim of fair representation learning is to learn an intermediate representation of the data that preserves as much information about the data as possible, while simultaneously removing protected characteristic information such as age and gender. 
Zemel et al. \cite{ZemWuSwePitetal13} learn a probabilistic mapping of the data point to a set of latent prototypes that is independent of protected characteristic (equality of acceptance rates, also called a statistical parity criterion), while retaining as much class label information as possible. 
Louizos et al. \cite{LouSweLiWeletal16} extend this by employing a deep variational auto-encoder (VAE) framework for finding the fair latent representation. 
In recent years, we see increased adversarial learning methods for fair representations. 
Ganin et al. \cite{GanUstAjaGeretal16} propose adversarial representation learning for domain adaptation by requiring the learned representation to be indiscriminate with respect to differences in the domains. 
Multiple data domains can be translated into multiple demographic groups.
Edwards and Storkey \cite{EdwSto16} make this connection and propose adversarial representation learning for the statistical parity criterion. 
To achieve other notions of fairness such as equality of opportunity, Beutel et al. \cite{BeuCheZhaChi17} show that the adversarial learning algorithm of Edwards and Storkey \cite{EdwSto16} can be reused but we only supply training data with positive outcome to the adversarial component.
Madras et al. \cite{MadCrePitZem18} use a label-aware adversary to learn fair and transferable latent representations for the statistical parity as well as equality of opportunity criteria.

\emph{None of the above} learn fair representations while simultaneously retaining the semantic meaning of the data. 
There is an orthogonal work on feature selection using human perception of fairness (e.g. \cite{GrgRedGumWel18}), while this approach undoubtedly retains the semantic meaning of tabular data, it has not been generalized to image data.
In an independent work to ours, Sattigeri et al. \cite{SatHofChe18} describe a similar motivation of producing fair representations in the input image domain; their focus is on creating a whole new image-like dataset, rather than conditioning on each input image.
%
%
Hence it is not possible to visualise a fair version for a given image as provided by our method (refer to Figures \ref{fig:faces} and \ref{fig:faces_attributes}). 
%

\section{Interpretability in Fairness by Residual Decomposition}
We will use the illustrative example of an automated job application screening system. 
Given input data (photographs, work experience, education and training, personal skills, etc.) $\x^n \in \Xcal$, output labels of performed well or not well $y^n \in \Ycal = \{+1, -1\}$, and protected characteristic values, such as \emph{race} or \emph{gender}, $s^n \in \{A,B,C,D,\ldots\}$, or \emph{age}, $s^n\in\mathbb{R}$, we would like to train a classifier $f$ that decides whether or not to invite a person for an interview.
We want the classifier to predict outcomes that are accurate with respect to $y^n$ but fair with respect to $s^n$.

\subsection{Fairness definitions}
\label{sec:fairnessdefinitions}
Much work has been done on mathematical definitions of fairness (e.g. \cite{KleMulRag16,Cho17}). 
It is widely accepted that no single definition of fairness applies in all cases, but will depend on the specific context and application of machine learning models \cite{WEF18}.
In this paper, we focus on the \emph{equality of opportunity} criterion that requires the classifier $f$ and the protected characteristic $s$ be independent, conditional on the label being positive
\footnote{With binary labels, it is assumed that positive label is a desirable/advantaged outcome, e.g. expected to perform well at the job.}, in shorthand notation $f\ci s\ |\ y = +1$.
Expressing the shorthand notation in terms of a conditional distribution, we have $\PP(f(\x)|s,y=+1) = \PP(f(\x)|y=+1)$.
With binary protected characteristic, this reads as equal true positive rates across the two groups, $\PP(f(\x)=+1|s=A,y=+1)=\PP(f(\x)=+1|s=B,y=+1)$.
Equivalently, the shorthand notation can also be expressed in terms of joint distributions, resulting in $\PP(f(\x),s|y=+1) = \PP(f(\x)|y=+1)\PP(s|y=+1)$.
The advantage of using the joint distribution expression is that the variable $s$ does not appear as a conditioning variable, making it straightforward to use the expression for a multi- or continuous value or even multiple protected characteristics.
%
%

\subsection{Residual decomposition}
We want to learn a data representation $\xtilde^n$ for each input $\x^n$ such that: a) it is able to predict the output label $y^n$, b) it protects $s^n$ according to a certain fairness criterion, c) it lies in the same space as $\x^n$, that is $\xtilde^n\in\Xcal$. 
The third requirement ensures the learned representation to have the same \emph{semantic meaning} as the input. 
For example, for images of people faces, the goal is to modify facial appearance in order to remove the protected characteristic information. 
For tabular data, we desire systematic changes in values of categorical features such as education (bachelors, masters, doctorate, etc.).  
Visualizing those systematic changes will give evidence on how our algorithm enforces a certain fairness criterion.
This will be a powerful tool, albeit all the powers hinge on \emph{observational data}, to scrutinize the interplay between fairness criterion, protected characteristics, and classification accuracy. 
%
We proceed by making the following decomposition assumption on $\x$:
\begin{align}
\label{eq:decomposition}
\phi(\x) =   \phi(\xtilde) + \phi(\xhat),
\end{align}
with $\xtilde$ to be the component that is independent of $s$, $\xhat$ denoting the component of $\x$ that is dependent on $s$, and $\phi(\cdot)$ is some \emph{pre-trained} feature map. 
We will discuss about the specific choice of this pre-trained feature map for both image and tabular data later in the section.
What we want is to learn a mapping from a source domain (input features) to a target domain (fair features with the semantics of the input domain), i.e.
$
T: \x \rightarrow \xtilde,  
$
and we will parameterize this mapping $T = T_{\omega}$ where $\omega$ is a class of autoencoding transformer network. 
For our architectural choice of transformer network, please refer to Section \ref{sec:experiments}.

To enforce the decomposition structure in \eq{eq:decomposition}, we need to satisfy two conditions: a) $\xtilde$ to be independent of $s$, and b) $\xhat$ to be dependent of $s$. 
Given a particular statistical dependence measure, the first condition can be achieved by \emph{minimizing} the dependence measure between $P = \{\phi(\xtilde^{1}),\ldots,\phi(\xtilde^{N})\} = \{\phi(T_{\omega}(\x^1)),\ldots,\phi(T_{\omega}(\x^N))\}$ and $S=\{s^1,\ldots,s^N\}$; $N$ is the number of training data points. 
For the second condition, we first define a \emph{residual}:
\begin{align}
\phi(\x) - \phi(\xtilde) = \phi(\x) - \phi(T_{\omega}(\x)) = \phi(\xhat),
\end{align}
where the last term is the data component that is \emph{dependent} on a protected characteristic $s$.
We can then enforce the second condition by \emph{maximizing} the dependence measure between $R = \{\phi(\xhat^{1}),\ldots,\phi(\xhat^{N})\} = \{\phi(\x^1) - \phi(T_{\omega}(\x^1)),\ldots,\phi(\x^N) - \phi(T_{\omega}(\x^N))\}$ and $S$. 
We use the decomposition property as a guiding mechanism to learn the parameters $\omega$ of the transformer network $T_{\omega}$.
%
%

In the fair and interpretable representation learning task, we believe using residual is well-motivated because we know that our generated fair features should be somewhat similar to our input features.
Residuals will make learning the transformer network easier. 
Taking into consideration that we do not have training data about the target fair features $\xtilde$, we should not desire the transformer network to take the input feature $\x$ and \emph{generate} a new output $\xtilde$.
Instead, it should just learn how to \emph{adjust} our input $\x$ to produce the desired output $\xtilde$.
The concept of residuals is universal, for example, a residual block has been used to speed up and to prevent over-fitting of a very deep neural network \cite{HeZhaRenSun16}, and a residual regression output has been used to perform causal inference in additive noise models \cite{MooJanPetSch09}.

Formally, given the $N$ training triplets $(X,S,Y)$, to find a fair and interpretable representation $\xtilde= T_{\omega}(\x)$, our optimization problem is given by:
\begin{align}
&  \mini_{T_\omega}   \underbrace{\sum_{n=1}^N\Lcal(T_{\omega}(\x^n),y^n)}_{\text{prediction loss}} + \lambda_1 \underbrace{\sum_{n=1}^{N}\|\x^n-T_{\omega}(\x^n)\|_2^2}_{\text{reconstruction loss}} + \nonumber\\
&+ \lambda_2\left(\underbrace{- \text{HSIC}(R,S|Y=+1) + \text{HSIC}(P,S|Y=+1)}_{\text{decomposition loss}}\right) 
\label{eq:optproblem}
\end{align} 
where $\text{HSIC}(\cdot,\cdot)$ is the statistical dependence measure, and $\lambda_i$ are trade-off parameters. 
HSIC is the Hilbert-Schmidt norm of the cross-covariance operator between reproducing kernel Hilbert spaces.
This is equivalent to a non-parametric distance measure of a joint distribution and the product of two marginal distributions using the Maximum Mean Discrepancy (MMD) criterion\cite{GreBorRasSchetal12}; MMD has been successfully used in fairnesss literature in it's own right \cite{LouSweLiWeletal16, QuaSha17}. Section \ref{sec:fairnessdefinitions} discusses defining statistical independence based on a joint distribution, contrasting this with a conditional distribution.
We use the biased estimator of HSIC \cite{GreBouSmoSch05,SonSmoGreBedetal12}: $
   \text{HSIC}_{\text{emp.}} = (N-1)^{-2}\tr HKHL,
   $
 where $K, L \in\mathbb{R}^{N\times N}$ are the kernel matrices for the \emph{residual}
 set $R$ and the protected characteristic set $S$ respectively, i.e.\ $K_{ij} = k(r^i, r^j)$ and
 $L_{ij} = l(s^i, s^j)$ (similar definition for measuring independence between sets $P$ and $S$). 
 We use a Gaussian RBF kernel function for both $k(\cdot,\cdot)$ and $l(\cdot,\cdot)$. 
 Moreover, $H_{ij}=\delta_{ij}-N^{-1}$ centres the observations of set $R$ and set $S$ in RKHS feature space. 
The prediction loss is defined using a softmax layer on the output of the transformer network. 
While in image data we add the total variation (TV) penalty \cite{MahVed15} on the fair representation to ensure spatial smoothness, we do not enforce any regularization term for tabular data. 
%
\noindent In summary, we learn a new representation $\xtilde{}$ that removes statistical dependence on the protected characteristic $s$ (by minimizing $\text{HSIC}(P,S|Y=+1)$) and enforces the dependence of the residual $\x-\xtilde$ and $s$ (by maximizing $\text{HSIC}(R,S|Y=+1)$). 
We can then train any classifier $f$ using this new representation, and it will inherently satisfy the fairness criterion \cite{MadCrePitZem18}.

\paragraph{Neural style transfer and pre-trained feature space}
Neural style transfer (e.g. \cite{GatEckBet15a,JohAlaFei2016}) is a popular approach to perform an image-to-image translation.
Our decomposition loss in \eq{eq:optproblem} is reminiscent of a style loss used in neural style transfer models.
The style loss is defined as the distance between second-order statistics of a style image and the translated image.
Excellent results \cite{GatEckBet15a,JohAlaFei2016,UlyLebVedLem16,ulyanov2017} on neural style transfer rely on pre-trained features.
Following this spirit, we also use a ``pre-trained'' feature mapping $\phi(\cdot)$ in defining our decomposition loss.
For image data, we take advantage of the powerful representation of deep convolutional neural networks (CNN) to define the mapping function \cite{GatEckBet15a}.
The feature maps of $\x$ in the layer $l$ of a CNN are denoted by $F^l_\x\in R^{N_l\times M_l}$ where $N_l$ is the number of the feature maps in the layer $l$ and $M_l$ is the height times the width of the feature map.
We use the vectorization of $F^l_\x$ as the required mapping $\phi(\x) = \text{vec}(F^l_\x)$.
Several layers of a CNN will be used to define the full mapping (see Section \ref{sec:experiments}). 
For tabular data, we use the following random Fourier feature \cite{RahRec08} mapping $\phi(\x) = \sqrt{\nicefrac{2}{D}}\ \text{cos}(\inner{\theta}{\x}+b)$ with a bias vector $b\in\RR^D$ that is uniformly sampled in $[0,2\pi]$, and a matrix 
$\theta\in\RR^{d\times D}$ where $\theta_{ij}$ is sampled from a Gaussian distribution. 
We have assumed the input data lies in a $d$-dimensional space, and we transform them to a $D$-dimensional space.

\section{Experiments}
\label{sec:experiments}

We gave an illustrative example about screening job applications, however, no such data is publicly available. 
We will instead use publicly available data to simulate the setting. 
We conduct the experiments using three datasets: the CelebA image dataset\footnote{\scriptsize\url{http://mmlab.ie.cuhk.edu.hk/projects/CelebA.html}} \cite{liu2015faceattributes}, the Diversity in Faces (DiF) dataset \footnote{\scriptsize\url{https://www.research.ibm.com/artificial-intelligence/trusted-ai/diversity-in-faces/}} \cite{DiF2019}, and the Adult income dataset\footnote{\scriptsize\url{https://archive.ics.uci.edu/ml/datasets/adult}} from the UCI repository \cite{Dua:2017}. 
The CelebA dataset has a total of $202,599$ celebrity images. The images are annotated with $40$ attributes that reflect appearance (hair color and style, face shape, makeup, for example), emotional state (smiling), gender, attractiveness, and age. For this dataset, we use gender as a binary protected characteristic, and attractiveness as the proxy measure of getting invited for a job interview in the world of fame. 
We randomly select $20$K images for testing and use the rest for training the model. 
The DiF dataset has only been introduced very recently and contains nearly a million human face images reflecting diversity in ethnicity, age and gender. 
We include preliminary results using 200K images for training and 200K images for testing our model on this dataset. The images are annotated with attributes such as race, gender and age (both continual and discretized into seven age groups) as well as facial landmarks and facial symmetry features. For this dataset, we use gender as a binary protected characteristic, and the discretized age groups as a predictive task. 
The Adult income dataset is frequently used to assess fairness methods. It comes from the Census bureau and the binary task is to predict whether or not an individual earns more than $\$ 50$K per year. It has a total of $45,222$ data instances, each with $14$ features such as gender, marital status, educational level, number of work hours per week. 
For this dataset, we follow \cite{ZemWuSwePitetal13} and consider gender as a binary protected characteristic. 
We use $28,222$ instances for training, and $15,000$ instances for testing. 
%
We enforce equality of opportunity as the fairness criteria throughout for the three experiments.

\begin{table*}[]
\centering
\scalebox{0.9}{%
\begin{tabular}{lcrccccccc}
\toprule
{} & \multicolumn{2}{c}{original $\x$} & \multicolumn{2}{c}{fair interpretable $\xtilde$} & \multicolumn{2}{c}{latent embedding $\mathbf{z}$} \\
{}  & Accuracy $\uparrow$ &  Eq. Opp $\downarrow$  & Accuracy $\uparrow$ &  Eq. Opp $\downarrow$ & Accuracy $\uparrow$ &  Eq. Opp $\downarrow$ \\
\midrule
1: \texttt{LR} &    $85.1\pm0.2$ &  $\mathbf{9.2\pm2.3}$  & $84.2\pm0.3$ &  $\mathbf{5.6\pm2.5}$  & $81.8\pm2.1$ &  $\mathbf{5.9\pm4.6}$ \\
2: \texttt{SVM}         &    $85.1\pm0.2$ &  $\mathbf{8.2\pm2.3}$  & $84.2\pm0.3$ &  $\mathbf{4.9\pm2.8}$  & $81.9\pm2.0$ &  $\mathbf{6.7\pm4.7}$ \\
3: \texttt{Fair Reduction LR}~\cite{Aga18} &    $85.1\pm0.2$ &  $\mathbf{14.9\pm1.3}$  & $84.1\pm0.3$ &  $\mathbf{6.5\pm3.2}$ & $81.8\pm2.1$ &  $\mathbf{5.6\pm4.8}$ \\
4: \texttt{Fair Reduction SVM}~\cite{Aga18} &    $85.1\pm0.2$ &  $\mathbf{8.2\pm2.3}$  & $84.2\pm0.3$ &  $\mathbf{4.9\pm2.8}$  & $81.9\pm2.0$ &  $\mathbf{6.7\pm4.7}$ \\
5: \texttt{Kamiran \& Calders LR}~\cite{KamCal12} &    $84.4\pm0.2$ &  $\mathbf{14.9\pm1.3}$  & $84.1\pm0.3$ &  $\mathbf{1.7\pm1.3}$  & $81.8\pm2.1$ &  $\mathbf{4.9\pm3.3}$ \\
6: \texttt{Kamiran \& Calders SVM}~\cite{KamCal12} &    $85.1\pm0.2$ &  $\mathbf{8.2\pm2.3}$  & $84.2\pm0.3$ &  $\mathbf{4.9\pm2.8}$  & $81.9\pm2.0$ &  $\mathbf{6.7\pm4.7}$ \\
7: \texttt{Zafar et al.}$^*$~\cite{ZafValRodGum17b} & $85.0\pm0.3$ &  $\mathbf{1.8\pm0.9}$  &  --- &    ---  &      --- &    --- \\
\bottomrule\\
\end{tabular}
}
\caption{Results of training multiple classifiers (rows 1--7) on $3$ different representations, $\x$, $\xtilde$, and $\mathbf{z}$. $\x$ is the original input representation, $\xtilde$ is the interpretable, fair representation introduced in this paper, and $\mathbf{z}$ is the latent embedding representation of Beutel et al. \cite{BeuCheZhaChi17}. We \textbf{boldface} Eq. Opp. since this is the fairness criterion (the lower the better). $^*$The solver of \texttt{Zafar et al.} fails to converge in 4 out of 10 repeats. Our learned representation $\xtilde$ achieves comparable fairness level to the latent representation $\mathbf{z}$, while maintaining the constraint of being in the same space as the original input.}
\label{table:benchmarking}
\end{table*}
\begin{figure*}[tb]
\begin{center}
\scalebox{0.95}{%
\begin{tabular}{c|c}
  \includegraphics[width=0.24\textwidth]{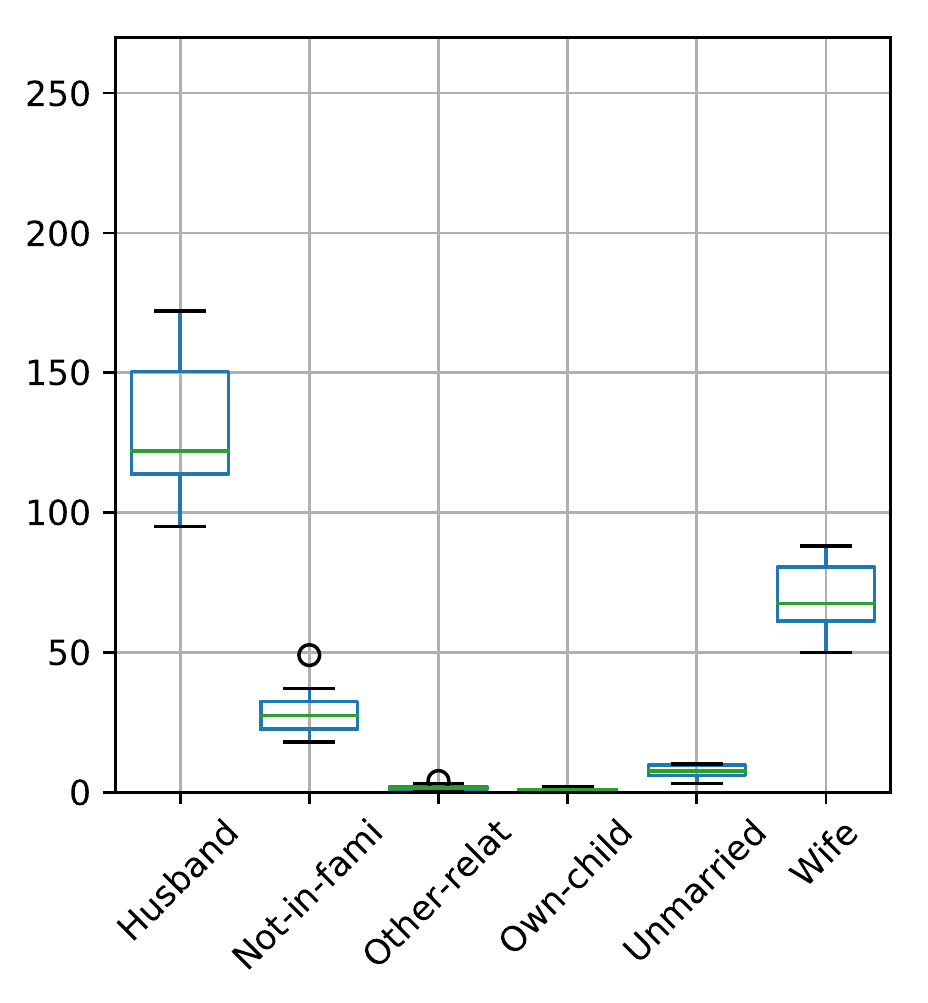}\includegraphics[width=0.24\textwidth]{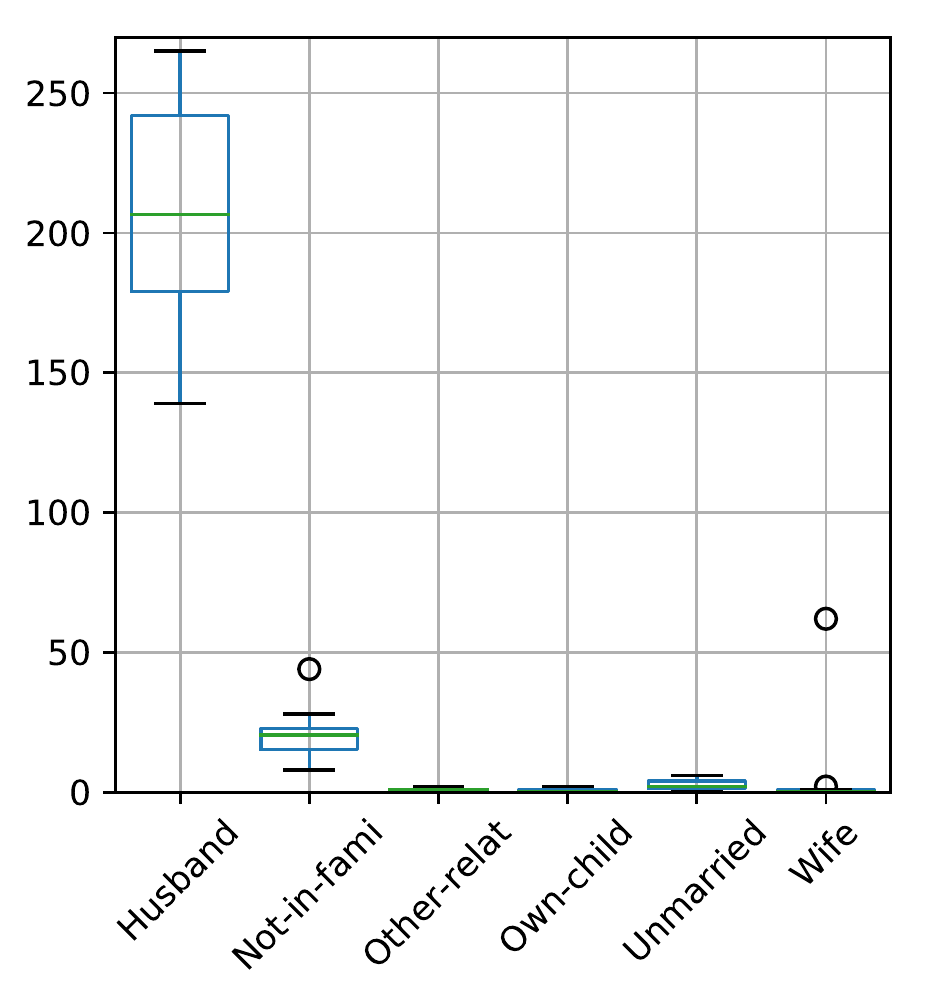}  & 
  \includegraphics[width=0.24\textwidth]{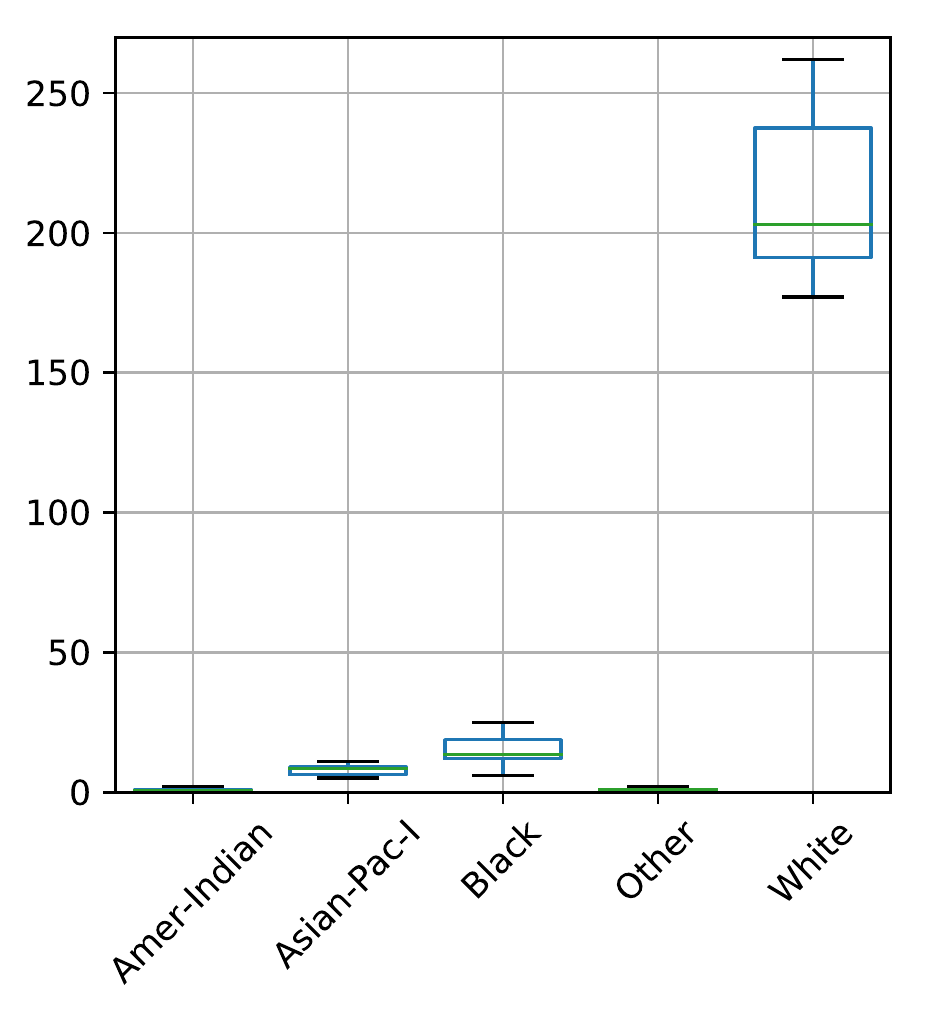}\includegraphics[width=0.24\textwidth]{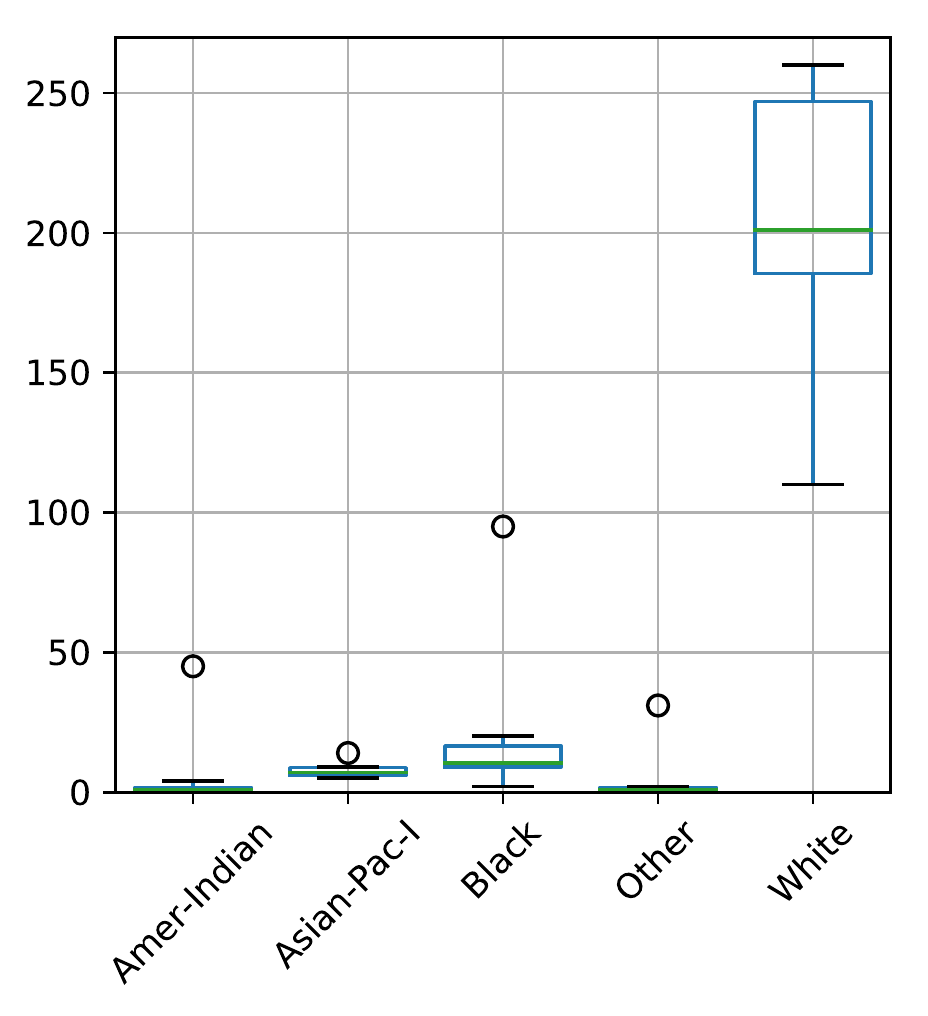}\\
  (`Relationship Status`) & (`Race`)
\end{tabular}
}
\caption{\textbf{Left} Boxplots showing the distribution of the categorical feature `Relationship Status` \textbf{Right} Boxplots showing the distribution of the categorical feature `Race`. 
\textbf{Left of each}: original representation $\x\in\Xcal$. \textbf{Right of each}: fair representation $\xtilde\in\Xcal$.\label{fig:interpretability}}
\end{center}
\end{figure*}

\subsection{The Adult Income dataset}
The focus is to investigate whether (\textbf{Q1}) our proposed fair and interpretable learning method performs on a par with state-of-the-art fairness methods, and whether (\textbf{Q2}) performing a tabular-to-tabular translation brings us closer to achieving interpretability in how fairness is being satisfied.
%
We compare our method against an unmodified $\x{}$ using the following classifiers: 
1) logistic regression (\texttt{LR}) and 
2) support vector machine with linear kernel (\texttt{SVM}),
We select the regularization parameter of \texttt{LR} and \texttt{SVM} over 6 possible values ($10^i$ for $i \in [0,6]$) using $3$-fold cross validation.
We then train classifiers 1--2 with the learned representation $\xtilde{}$ and with the latent embedding $\z{}$ of a state-of-the-art adversarial model described in Beutel et al. \cite{BeuCheZhaChi17}. 
We also apply methods which reweigh the samples to simulate a balanced dataset with regard to the protected characteristic FairLearn \cite{Aga18} \texttt{Fair Reduction} 3-4 and Kamiran \& Calders \cite{KamCal12} \texttt{Kamiran \& Calders} 5-6, 
optimized with both the cross-validated \texttt{LR} and \texttt{SVM} (1-2), 
giving (\texttt{Fair Reduction LR}), (\texttt{Fair Reduction SVM}), (\texttt{Kamiran \& Calders LR}) and (\texttt{Kamiran \& Calders SVM}) respectively.
As a reference, we also compare with:
7) Zafar et al.'s~\cite{ZafValRodGum17b} fair classification method (\texttt{Zafar et al.}) that adds equality of opportunity directly as a constraint to the learning objective function.
It has been shown that applying fairness constraints in succession as `fair pipelines' do not enforce fairness \cite{DwoIlv18, BowKitNisStraVarVen17}, as such, we only demonstrate (fair) classifier 7 on the unmodified $\x{}$.

\textbf{Benchmarking}
We train our model for $50,000$ iterations using a network with 1 hidden layer of $40$ nodes for both the encoder and decoder, with the encoded representation being 40 nodes. The predictor acts on the decoded output of this network. We set the trade-off parameters of the reconstruction loss ($\lambda_1$) and decomposition loss ($\lambda_2$) to $10^{-4}$ and $100$ respectively.
We then use this model to translate $10$ different training and test sets into $\xtilde{}$.
Using a modified version of the framework provided by Friedler et al. \cite{FrieSchVen18} we evaluate methods $1$--$6$ using $\x{}$ and $\xtilde{}$ representations. To ensure consistency, we train the model of Beutel et al. \cite{BeuCheZhaChi17} with the same architecture and number of iterations as our model.

Table \ref{table:benchmarking} shows the results of these experiments. Our interpretable representation, $\xtilde{}$ achieves similar fairness level to Beutel's state-of-the-art approach (\textbf{Q1}). Consistently, our representation $\xtilde{}$ promoted the \emph{fairness} criterion (Eq. Opp. close to $0$), with only a small penalty in accuracy.

\textbf{Interpretability}
We promote equality of opportunity for the positive class ($\text{actual salary} > \text{\$50K}$). 
In Figure \ref{fig:interpretability} we show the effect of learning a fair representation, showing changes in the `Relationship Status' and `Race' features of samples that were incorrectly classified by an SVM as earning $<$ \$50K in $\x{}$, but were correctly classified in $\xtilde{}$. 
%
%
The visualization can be used for understanding how representation methods adjust the data for fairness. 
For example in Figure \ref{fig:interpretability} (left) we can see that our method deals with the notorious problem of a husband or wife relationship status being a direct proxy for gender (\textbf{Q2}). Our method recognises this across all repeats in an unsupervised manner and reduces the wife category which is associated with a negative prediction. Other categories that have less correlation with the protected characteristic, such as race, largely remain unmodified (Figure \ref{fig:interpretability} (right)). 

\subsection{The CelebA dataset}
Our intention here is to investigate whether (\textbf{Q3}) performing an image-to-image translation brings us closer to achieving interpretability in how fairness is being satisfied, and whether (\textbf{Q4}) using semantic attribute representations of images reinforces similar interpretability conclusions as using image features directly.

\begin{figure}[t]
\begin{tabular}{l}
\hspace{-0.2cm}translated\hspace{-0.4cm}
\end{tabular}
\begin{tabular}{l}
\includegraphics[width=0.095\textwidth]{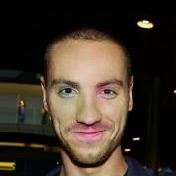} 
\includegraphics[width=0.095\textwidth]{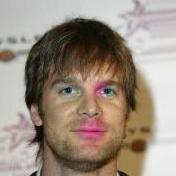} 
\includegraphics[width=0.095\textwidth]{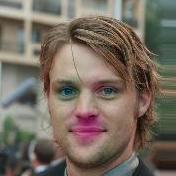} 
\includegraphics[width=0.095\textwidth]{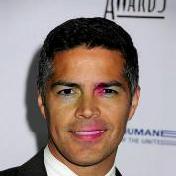} 
\end{tabular}\\
\begin{tabular}{l}
\hspace{-0.2cm}residual\hspace{-0.15cm}  
\end{tabular}
\begin{tabular}{l}
\includegraphics[width=0.095\textwidth]{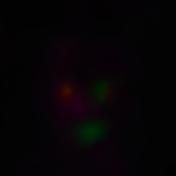}
\includegraphics[width=0.095\textwidth]{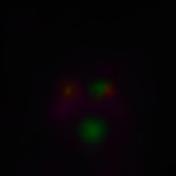}
\includegraphics[width=0.095\textwidth]{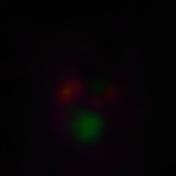}
\includegraphics[width=0.095\textwidth]{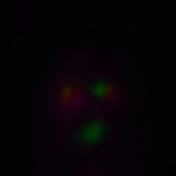}
\end{tabular}
\caption{Examples of the translated and residual images on CelebA from the protected group of males (minority group) that have been classified correctly (as attractive) after transformation. These results are obtained with the transformer network for image-to-image translation. 
Best viewed in color.}
\label{fig:faces}
\end{figure}
\begin{table}[tb!]
  \centering\scalebox{0.85}{
  \begin{tabular}{lccccc}
    \multicolumn{6}{c}{}\\
    \toprule
    & domain     & Acc.  & Eq. Opp.     &  TPR & TPR \\
    &  $\Xcal$                 & $\uparrow$& $\downarrow$ & ${\text{female}}$ & ${\text{male}}$\\
    \midrule
    1: orig. $\x $ & \emph{images}           & $80.6$   & $\mathbf{33.8}$ & $90.8$ & $57.0$\\
    2: orig. $\x $ & \emph{attributes}       & $79.1$   & $\mathbf{39.9}$ & $90.8$ & $50.9$\\    
    3: fair $\xtilde$ & \emph{images}           & \\
     \multicolumn{2}{l}{a: $\lambda_2=1.00$, $\text{biased}\ \text{HSIC}$} & $79.4$   & $\mathbf{23.8}$ & $85.2$ & $61.4$\\
    \multicolumn{2}{l}{b: $\lambda_2=10.0$, $\text{biased}\ \text{HSIC}$} & $80.3$   & $\mathbf{22.8}$ & $85.6$ & $62.7$\\
    \multicolumn{2}{l}{c: $\lambda_2=10.0$, $\text{unbiased}\ \text{HSIC}$} & $80.2$ & $\mathbf{18.7}$ & $84.3$ & $65.6$\\
    4: fair $\xtilde$ & \emph{attributes}       & $75.9$   & $\mathbf{12.4}$ & $87.2$ & $74.8$\\
    5: fair $\xtilde$ & \emph{fake images}      & $78.5$   & $\mathbf{23.0}$ & $87.5$ & $64.5$\\
    \bottomrule\\
  \end{tabular}}
\caption{Results on CelebA dataset using a variety of input domains. 
    Prediction performance is measured by accuracy, and we use equality of opportunity, TPRs difference, as the fairness criterion. 
    Here, domain of fake images (last row) denotes images synthesized by the StarGAN\cite{ChoChoKimHaetal18} model from the original images and their fair attribute representations. We \textbf{boldface} Eq. Opp. since this is the fairness criterion.
    }
      \label{tab:results_celeba}
\end{table}

\textbf{Image-to-image translation} 
Our autoencoder network is based on the architecture of the transformer network for neural style transfer \cite{JohAlaFei2016} with three convolutional layers, five residual layers and three deconvolutional/upsampling layers in combination with instance weight normalization \cite{ulyanov2017}. 
The transformer network produces the residual image using a non-linear tanh activation, which is then subtracted from the input image to form the translated fair image $\xtilde$.
%
Similarly to neural style transfer \cite{GatEckBet15a, gardner2016, JohAlaFei2016}, for computing the loss terms, we use the activations in the deeper layers of the 19-layered VGG19 network \cite{SimZis15} as feature representations of both input and translated images. Specifically, we use activations in the conv3\_1, conv4\_1 and conv5\_1 layers for computing the decomposition loss, the conv3\_1 layer activations for the reconstruction loss, and the activations in the last convolutional layer pool\_5 for the prediction loss and when evaluating the performance. Given a 176x176 color input image, we compute the activations at each layer mentioned earlier after ReLU, then we flatten and $l_2$ normalize them to form features for the loss terms.
In the HSIC estimates of the decomposition loss, we use a Gaussian RBF kernel $k(x_1,x_2) = \text{exp} (-\gamma \|x_1-x_2\|^2)$ width $\gamma=1.0$ for image features, and $\gamma =0.5$ for protected characteristics (as one over squared distance in the binary space).
To compute the decomposition loss, we add the contributions across the three feature layers. 
%
We set the trade-off parameters $\lambda_1$ and $\lambda_2$ of the reconstruction loss and the decomposition loss, respectively, to $1.0$\footnote{We also try $\lambda_2=10.0$ to further enforce equality of opportunity, and use the unbiased estimator of HSIC \cite{SonSmoGreBedetal12}.}, and the TV regularization strength 
to $10^{-3}$. 
Training was carried out for 50 epochs with a batch size of $80$ images. 
We use minibatch SGD and apply the Adam solver \cite{kingma2014adam} with learning rate $10^{-3}$; our TensorFlow implementation is publicly available\footnote{\url{https://github.com/predictive-analytics-lab/Data-Domain-Fairness}}. 

\textbf{Benchmarking and interpretability} 
We enforce equality of opportunity as the fairness criterion, and we consider attractiveness as the positive label. 
Attractiveness is what could give someone a job opportunity or an advantaged outcome as defined in \cite{HarPriSre16}. 
To test the hypothesis that we have learned a fairer image representation, we compare the performance and fairness of a standard SVM classifier trained using original images and the translated fair images. We use activation in the pool\_5 layer of the VGG19 network as features for training and evaluating the classifier\footnote{We deliberately evaluate the performance (accuracy and fairness) using an auxiliary classifier instead of using the predictor of the transformer network. Since the emphasis of this work is on representation learning, we should not prescribe what classifier the user chooses on top of learned representation.}.

We report the quantitative results of this experiment in Table \ref{tab:results_celeba} (first and third rows) and the qualitative evaluations of image-to-image translations in Figure \ref{fig:faces}.  
From the Table \ref{tab:results_celeba} it is clear that the classifier trained on fair/translated images $\xtilde$ has improved over the classifier trained on the original images $\x$ in terms of equality of opportunity (reduction from $33.8$ to $23.8$) while maintaining the prediction accuracy ($79.4$ comparing to $80.6$). 
The reduction in equality of opportunity can be further improved by increasing the parameter $\lambda_2$ to $10.0$ (third row (b)), and by using unbiased estimator of HSIC (third row (c)).  
Looking at the TPR values across protected features (females and males), we can see that the male TPR value has increased, but it has an opposite effect for females. 
In the CelebA dataset, the proportion of attractive to unattractive males is around $30\%$ to $70\%$, and it is opposite for females; male group is therefore the minority group in this problem. 
Our method achieves better equality of opportunity measure than the baseline by increasing the minority group TPR value while decreasing the majority group TPR value.
To understand the balancing mechanism of TPR values (\textbf{Q3}), we visualize a subset of test male images that have been classified correctly as attractive after transformation (those examples were misclassified in the original domain) in Figure \ref{fig:faces}. 

\begin{figure}[t]
\begin{tabular}{ll}
\begin{tabular}{l}
\hspace{-0.4cm}input
\end{tabular}
\begin{tabular}{l}
\hspace{-0.32cm}
\includegraphics[width=0.08\textwidth]{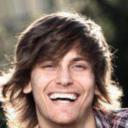}
\includegraphics[width=0.08\textwidth]{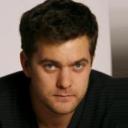}
\includegraphics[width=0.08\textwidth]{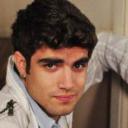}
\includegraphics[width=0.08\textwidth]{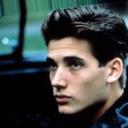} 
\includegraphics[width=0.08\textwidth]{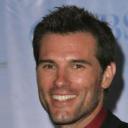} 
\end{tabular}\\
\begin{tabular}{l}
\hspace{-0.4cm}output 
\end{tabular}
\begin{tabular}{l}
\hspace{-0.5cm}
\includegraphics[width=0.08\textwidth]{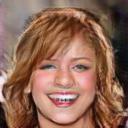}
\includegraphics[width=0.08\textwidth]{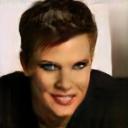}
\includegraphics[width=0.08\textwidth]{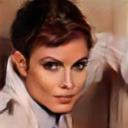}
\includegraphics[width=0.08\textwidth]{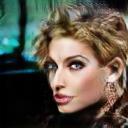}
\includegraphics[width=0.08\textwidth]{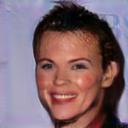}
\end{tabular}
\end{tabular}
\caption{Results of our approach (image-to-image translation via attributes). Given $N$ i.i.d. samples $\{(\x^n,y^n)\}_{n=1}^N$, our method transforms them into a new fair dataset $\{(\tilde{\x}^n,y^n)\}_{n=1}^{N}$ where $(\tilde{\x}^n,y^n)$ is the fair version of $(\x^n,y^n)$. 
The synthesized images are produced by the StarGAN model \cite{ChoChoKimHaetal18} conditioned on the original images and their fair attribute representation. }
\label{fig:faces_attributes}
\end{figure}
\begin{figure}[t!b]
\centering
   \includegraphics[width=0.23\textwidth]{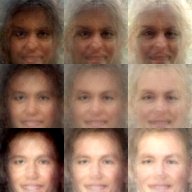}
   \includegraphics[width=0.23\textwidth]{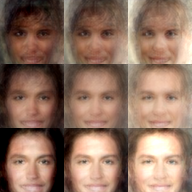}
   \caption{Results of Fainess GAN \cite{SatHofChe18} (Fig.2) of non-attractive (left) and attractive (right) males after pre-processing. 
   Given $N$ i.i.d. samples $\{(\x^n,y^n)\}_{n=1}^N$, Fainess GAN transforms them into a new fair dataset $\{(\tilde{\x}^n,\tilde{y}^n)\}_{n=1}^{N'}$ where $N'\neq N$ and $(\tilde{\x}^n,\tilde{y}^n)$ has no correspondence to $(\x^n,y^n)$.}
\label{fig:fairgan}
\end{figure}
\begin{figure}[t!b]
\centering
\includegraphics[width=0.38\textwidth]{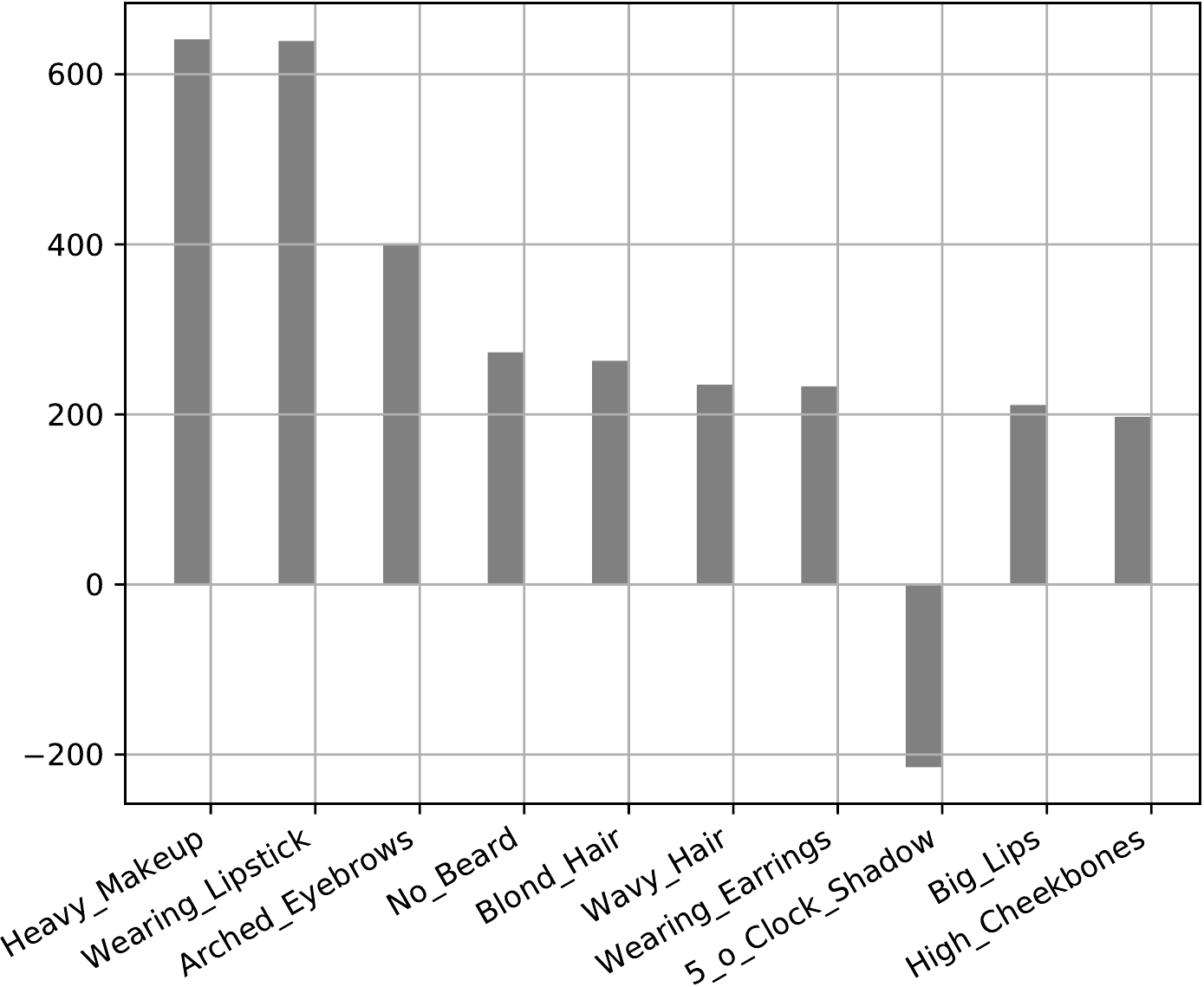}
\caption{Top 10 semantic attribute features that have been changed in $647$ males; those males were \emph{incorrectly} predicted as not attractive, but are now correctly predicted as attractive. $641$ and $639$ males out of $647$ are now with ``Heavy\_Makeup'' and ``Wearing\_Lipstick'' attributes, respectively, and $215$ out of $647$ males are now \emph{without} a ``5\_o\_Clock\_Shadow'' attribute.}  
\label{fig:attributes}
\end{figure}
We observe a consistent localized area in face, specifically \emph{lips} and \emph{eyes} regions. 
The CelebA dataset has a large diversity in visual appearance of females and males (hair style, hair color) and their ethnic groups, so more localized facial areas have to be discovered to equalize TPR values across groups. 
Lips are very often coloured in female (the majority group) celebrity faces, hence our method, to increase the minority group TPR value, colorizes the lip regions of the minority group (males). 
Interestingly, female faces without prominent lipstick often got this transformation as well, prompting the decrease in the majority group TPR value. 
Regarding eye regions, several studies (e.g. \cite{brown1993} and references therein) have shown their importance in gender identification. 
Also, a heavy makeup that is often applied to female celebrity eyes can also support our visualization in Figure \ref{fig:faces}.

The image-to-image translation using transformer network learns to produce coarse-grained changes, i.e. masking/colorizing face regions. 
This is expected as we learn a highly unconstrained mapping from source to target domain, in which the target data is unavailable. 
To enable fine-grained changes and semantic transformation of the images, we now explore semantic attributes; attributes are well-established interpretable mid-level representations for images. 
We show how an attribute-to-attribute translation provides an alternative way in analysing and performing an image-to-image translation. 

\textbf{Attribute-to-attribute translation}
Images in the CelebA dataset come with $40$ dimensional binary attribute annotations. We use all but two attributes (\emph{gender} and \emph{attractiveness}) as semantic attribute representation of images. 
We then perform attribute-to-attribute translation with the transformer network and consider the same attractive versus not attractive task and gender protected characteristic as with the image data. 
We report the results of this experiment in Table \ref{tab:results_celeba} (second and forth rows correspond to the domain of attributes).
First, we observe that the predictive performance of the classifier trained on attribute representation is only slightly lower than the performance of the classifier trained on the image data ($79.1$ versus $80.6$), which enables sensible comparison of the results in these two settings. 
Second, we observe better gain in equality of opportunity when using the transformed attribute representation comparing to transformed images ($12.4$ is the best Eq. Opp. result in this experiment). 
This comes at the cost of a drop in accuracy performance. 
The TPR rates for both groups are higher when using translated attribute representation than when using translated image representation (third row versus fourth row). 
The largest improvement of the TPR is observed in the group of males (from $50.9$ in the original attribute to $74.8$ in the translated attribute space). 
Further analysis of changes in attribute representation reveals that equality of opportunity is achieved by putting \emph{lipstick} and \emph{heavy-makeup} to the male group (Figure \ref{fig:attributes}). 
These top 2 features have been mostly changed in the group of \emph{males}. 
Very few changes happened in the group of females. 
This is encouraging as we have just arrived at the same conclusion (Figures \ref{fig:faces} and \ref{fig:attributes}), be it using images or using semantic attributes (\textbf{Q4}).

\textbf{Image-to-image translation via attributes}
Given the remarkable progress that has been made in the field towards image synthesis with the conditional GAN models, we attempt to synthesize images with respect to the attribute description.  
Specifically, we use the StarGAN model \cite{ChoChoKimHaetal18}, the state-of-the-art model for image synthesis with multi-attribute transformation, to synthesize images with our learned fair attribute representation. 
For this, we pre-train the StarGAN model to perform image transformations with $38$ binary attributes (excluding gender and attractive attributes) using training data. 
We then translate all images in CelebA with respect to their fair attribute representation.  
We evaluate the performance of this approach and report the results in Table \ref{tab:results_celeba} (last row). 
We also include the qualitative evaluations of image-to-image translations via attributes in Figure \ref{fig:faces_attributes}. 
These visualizations essentially generalize counterfactual explanations in the sense of \cite{WacMitRus18} to the image domain. 
We have just shown the ``closest synthesized world'', i.e. the smallest change to the world that can be made to obtain a desirable outcome.
Overall, the classifier trained using this fair representation shows similar Eq. Opp. performance and comparable accuracy to the classifier trained on representation learned with the transformer network. 
However, the TPR rates for both protected groups are higher (last row versus third row), especially in the group of males, when using this representation. 
%

\textbf{Pre-processing approaches}
The aim of the pre-processing approaches such as \cite{SatHofChe18, calmon2017optimized} is to transform the given dataset of $N$ i.i.d. samples $\{(\x^n,y^n)\}_{n=1}^N$ into a \emph{new} fair dataset $\{(\tilde{\x}^n,\tilde{y}^n)\}_{n=1}^{N'}$. 
It is important to note that $N'$ is not necessarily equal to $N$, and therefore $(\tilde{\x}^n,\bar{y}^n)$ has no correspondence to $(\x^n,y^n)$. 
\cite{calmon2017optimized} has proposed this approach for tabular (discrete) data, while \cite{SatHofChe18} has explored image data. 
Here, we offer a unified framework for tabular (continuous and discrete) and image data that transforms the given dataset $\{(\x^n,y^n)\}_{n=1}^N$ into a new fair dataset $\{(\tilde{\x}^n,y^n)\}_{n=1}^{N}$ where $(\tilde{\x}^n,y^n)$ is the fair version of $(\x^n,y^n)$. 
\emph{What is the advantage of creating a fair representation per sample (our method) rather than on the whole dataset at once \cite{SatHofChe18, calmon2017optimized}?}
The first can be used to provide an \emph{individual}-level explanation of fair systems, while the latter can only be used to provide a \emph{system}-level explanation.
For comparison, we include here a snapshot of results presented in \cite{SatHofChe18} using the CelebA dataset in Figure \ref{fig:fairgan}. 
The figure shows eigenfaces/eigensketches with \emph{the mean image} of the new fair dataset $\{(\tilde{\x}^n)\}_{n=1}^{N'}$ (in the center) of the $3\times 3$ grid.
No per sample visualisation $(\tilde{\x}^n)$ was provided.
Left/right/top/bottom images in Fig. \ref{fig:fairgan} show variations along the first/second principal components. 
In contrast, Figure \ref{fig:faces_attributes} shows a per sample visualisation $(\tilde{\x}^n)$ using our proposed method.

\subsection{The Diversity in Faces dataset}
We extract and align face crops from the images and use 128x128 facial images as the inputs. 
Our preliminary experiment has similar setup to the image-to-image translation on the CelebA dataset except that the prediction task has seven age groups to be classified.  
As the fairness criterion we enforce equality of opportunity considering the middle age group (31-45) to be desirable (as the positive label when conditioning). 
As before, to test the hypothesis that we have learned a fairer image representation, we compare the performance and fairness of the SVM classifier trained using original images and the translated fair images (with features as activations in the pool\_5 layer of the VGG19 network). 
%
We achieve $52.85$ as the overall classification accuracy over seven age groups when using original image features and an increased $60.26$ accuracy when using translated images. 
The equality of opportunity improved from $27.21$ using original image representation to $9.85$ using fair image representation. 
%
%
Similarly to the CelebA dataset, the image-to-image translation using transformer network learns to produce coarse-grained changes, i.e. masking/colorizing nose regions (as opposed to lips and eyes regions on CelebA). 
These preliminary results are encouraging and further analysis will be addressed as a future extension.  

\section{Discussion and Conclusion}

\begin{quotation}
{\flushright\textit{It is not clear if fairness and interpretability are conflicting requirements.}\vspace{-0.45cm}}
\flushright{Reviewer \#1}\vspace{-0.05cm}
\end{quotation}

%
They are not, however interpretability in how fairness is enforced has so far been overlooked despite 
being an integral ingredient for encouraging productive public debates regarding fair machine learning systems.
%
Interpretability in machine learning models can help to ascertain qualitatively \emph{whether} fairness is met \cite{RS2017,DosKim17}. 
This paper takes a step further and advocates interpretability to ascertain qualitatively \emph{how} fairness is met, once we have agreed to 
enforce fairness (e.g. equality of opportunity) in  machine learning models.  
We specifically focus on enforcing fairness in representation learning.
Unlike other fair representation learning methods that learn \emph{latent} embeddings, our method learns a representation that is in the same space as the original input data, therefore retaining the semantics of the input domain.
Our method picks up consistently in \emph{$10$ out of $10$ repeated experiments} whether a person is a husband or wife as a direct proxy for \emph{gender}, and subsequently reduces the wife category which is associated with a negative prediction.
In our experiments with people's faces, eyes and lips are considered to be the direct proxy for gender attractiveness, and nose regions for being in a certain age group.
As a potential future direction, we plan to further analyze the interpretability in fairness using causal reasoning \cite{LopNisChi17}. 

\section*{Acknowledgments}
\noindent NQ is supported by the UK EPSRC project EP/P03442X/1 and the Russian Academic Excellence Project` 5-100'. 
VS is supported by the Imperial College Research Fellowship.
We gratefully acknowledge NVIDIA for GPUs donation and Amazon for AWS Cloud Credits.

{\small
\bibliographystyle{ieee_fullname}
\bibliography{bibfile}

\begin{thebibliography}{10}\itemsep=-1pt

\bibitem{Aga18}
Alekh Agarwal, Alina Beygelzimer, Miroslav Dudik, John Langford, and Hanna
  Wallach.
\newblock A reductions approach to fair classification.
\newblock In {\em International Conference on Machine Learning (ICML)}, 2018.

\bibitem{BeuCheZhaChi17}
Alex Beutel, Jilin Chen, Zhe Zhao, and Ed~H. Chi.
\newblock Data decisions and theoretical implications when adversarially
  learning fair representations.
\newblock {\em CoRR}, abs/1707.00075, 2017.

\bibitem{BowKitNisStraVarVen17}
Amanda Bower, Sarah~N. Kitchen, Laura Niss, Martin~J. Strauss, Alexander
  Vargas, and Suresh Venkatasubramanian.
\newblock Fair pipelines.
\newblock {\em CoRR}, abs/1707.00391, 2017.

\bibitem{brown1993}
Elizabeth Brown and David~I Perrett.
\newblock What gives a face its gender?
\newblock {\em Perception}, 22(7):829--840, 1993.

\bibitem{BruBurHan93}
V. Bruce, A. Burton, E. Hanna, P. Healey, O. Mason, A. Coombes, R. Fright, and
  A. Linney.
\newblock Sex discrimination: how do we tell the difference between male and
  female faces?
\newblock {\em Perception}, 22, 1993.

\bibitem{calmon2017optimized}
Flavio Calmon, Dennis Wei, Bhanukiran Vinzamuri, Karthikeyan~Natesan
  Ramamurthy, and Kush~R Varshney.
\newblock Optimized pre-processing for discrimination prevention.
\newblock In {\em Advances in Neural Information Processing Systems (NIPS)},
  pages 3992--4001, 2017.

\bibitem{ChoChoKimHaetal18}
Yunjey Choi, Minje Choi, Munyoung Kim, Jung-Woo Ha, Sunghun Kim, and Jaegul
  Choo.
\newblock Stargan: Unified generative adversarial networks for multi-domain
  image-to-image translation.
\newblock In {\em Computer Vision and Pattern Recognition (CVPR)}, 2018.

\bibitem{Cho17}
A. Chouldechova.
\newblock Fair prediction with disparate impact: A study of bias in recidivism
  prediction instruments.
\newblock {\em Big data}, 5, 2017.

\bibitem{Dua:2017}
Dua Dheeru and Efi Karra~Taniskidou.
\newblock {UCI} machine learning repository, 2017.

\bibitem{DosKim17}
Finale Doshi{-}Velez and Been Kim.
\newblock Towards a rigorous science of interpretable machine learning.
\newblock {\em CoRR}, abs/1702.08608, 2017.

\bibitem{DwoIlv18}
Cynthia Dwork and Christina Ilvento.
\newblock Fairness under composition.
\newblock In {\em Innovations in Theoretical Computer Science (ITCS)}, 2019.

\bibitem{EdwSto16}
Harrison Edwards and Amos Storkey.
\newblock Censoring representations with an adversary.
\newblock In {\em International Conference in Learning Representations (ICLR)},
  2016.

\bibitem{FrieSchVen18}
Sorelle~A. Friedler, Carlos Scheidegger, Suresh Venkatasubramanian, Sonam
  Choudhary, Evan~P. Hamilton, and Derek Roth.
\newblock A comparative study of fairness-enhancing interventions in machine
  learning.
\newblock {\em CoRR}, abs/1802.04422, 2018.

\bibitem{FuHeHou2014}
Siyao Fu, Haibo He, and Zeng-Guang Hou.
\newblock Learning race from face: A survey.
\newblock {\em IEEE Transactions on Pattern Analysis and Machine Intelligence
  (TPAMI)}, pages 2483--2509, 2014.

\bibitem{GanUstAjaGeretal16}
Yaroslav Ganin, Evgeniya Ustinova, Hana Ajakan, Pascal Germain, Hugo
  Larochelle, Fran\c{c}ois Laviolette, Mario Marchand, and Victor Lempitsky.
\newblock Domain-adversarial training of neural networks.
\newblock {\em Journal of Machine Learning Research (JMLR)}, 17:2096--2030,
  2016.

\bibitem{gardner2016}
Jacob~R Gardner, Paul Upchurch, Matt~J Kusner, Yixuan Li, Kilian~Q Weinberger,
  Kavita Bala, and John~E Hopcroft.
\newblock Deep manifold traversal: Changing labels with convolutional features.
\newblock In {\em European Conference on Computer Vision (ECCV)}, 2016.

\bibitem{GatEckBet15a}
Leon~A. Gatys, Alexander~S. Ecker, and Matthias Bethge.
\newblock A neural algorithm of artistic style.
\newblock {\em CoRR}, abs/1508.06576, 2015.

\bibitem{WEF18}
{Global Future Council on Human Rights 2016-18}.
\newblock How to prevent discriminatory outcomes in machine learning.
\newblock Technical report, World Economic Forum, 2018.

\bibitem{GreBorRasSchetal12}
Arthur Gretton, Karsten~M. Borgwardt, Malte~J. Rasch, Bernhard Sch\"{o}lkopf,
  and Alexander Smola.
\newblock A kernel two-sample test.
\newblock {\em Journal of Machine Learning Research (JMLR)}, 13:723--773, 2012.

\bibitem{GreBouSmoSch05}
Arthur Gretton, Olivier Bousquet, Alexander~J. Smola, and Bernhard
  Sch{\"{o}}lkopf.
\newblock Measuring statistical dependence with {Hilbert-Schmidt} norms.
\newblock In {\em Algorithmic Learning Theory {ALT}}, 2005.

\bibitem{GrgRedGumWel18}
N. Grgic-Hlaca, E. Redmiles, K.~P. Gummadi, and A. Weller.
\newblock Human perceptions of fairness in algorithmic decision making: A case
  study of criminal risk prediction.
\newblock In {\em The Web Conference (WWW)}, 2018.

\bibitem{HarPriSre16}
Moritz Hardt, Eric Price, and Nati Srebro.
\newblock Equality of opportunity in supervised learning.
\newblock In {\em Advances in Neural Information Processing Systems (NIPS) 29},
  2016.

\bibitem{HeZhaRenSun16}
Kaiming He, Xiangyu Zhang, Shaoqing Ren, and Jian Sun.
\newblock Deep residual learning for image recognition.
\newblock In {\em Computer Vision and Pattern Recognition (CVPR)}, 2016.

\bibitem{IsoZhuZhoEfr17}
P. Isola, J.-Y. Zhu, T. Zhou, and A.~A. Efros.
\newblock Image-to-image translation with conditional adversarial networks.
\newblock In {\em Computer Vision and Pattern Recognition (CVPR)}, 2017.

\bibitem{JohAlaFei2016}
Justin Johnson, Alexandre Alahi, and Li Fei-Fei.
\newblock Perceptual losses for real-time style transfer and super-resolution.
\newblock In {\em European Conference on Computer Vision (ECCV)}, 2016.

\bibitem{KamCal12}
Faisal Kamiran and Toon Calders.
\newblock Data preprocessing techniques for classification without
  discrimination.
\newblock {\em Knowledge and Information Systems}, 33:1--33, 2012.

\bibitem{kingma2014adam}
Diederik~P Kingma and Jimmy Ba.
\newblock Adam: A method for stochastic optimization.
\newblock In {\em International Conference on Learning Representations (ICLR)},
  2014.

\bibitem{KleMulRag16}
Jon~M. Kleinberg, Sendhil Mullainathan, and Manish Raghavan.
\newblock Inherent trade-offs in the fair determination of risk scores.
\newblock {\em CoRR}, abs/1609.05807, 2016.

\bibitem{LevHas15}
Gil Levi and Tal Hassner.
\newblock Age and gender classification using convolutional neural networks.
\newblock In {\em Computer Vision and Pattern Recognition Workshops (CVPRW)},
  2015.

\bibitem{liu2015faceattributes}
Ziwei Liu, Ping Luo, Xiaogang Wang, and Xiaoou Tang.
\newblock Deep learning face attributes in the wild.
\newblock In {\em International Conference on Computer Vision (ICCV)}, 2015.

\bibitem{LopNisChi17}
David Lopez-Paz, Robert Nishihara, Soumith Chintala, Bernhard Sch\"{o}lkopf,
  and L\'{e}on Bottou.
\newblock Discovering causal signals in images.
\newblock In {\em Computer Vision and Pattern Recognition (CVPR)}, 2017.

\bibitem{LouSweLiWeletal16}
Christos Louizos, Kevin Swersky, Yujia Li, Max Welling, and Richard Zemel.
\newblock The variational fair autoencoder.
\newblock In {\em International Conference on Learning Representations (ICLR)},
  2016.

\bibitem{MadCrePitZem18}
David Madras, Elliot Creager, Toniann Pitassi, and Richard~S. Zemel.
\newblock Learning adversarially fair and transferable representations.
\newblock In {\em International Conference on Machine Learning (ICML)}, 2018.

\bibitem{MahVed15}
Aravindh Mahendran and Andrea Vedaldi.
\newblock Understanding deep image representations by inverting them.
\newblock In {\em Computer Vision and Pattern Recognition (CVPR)}, 2015.

\bibitem{DiF2019}
Michele Merler, Nalini~K. Ratha, Rog{\'{e}}rio~Schmidt Feris, and John~R.
  Smith.
\newblock Diversity in faces.
\newblock {\em CoRR}, abs/1901.10436, 2019.

\bibitem{MirOsi14}
Mehdi Mirza and Simon Osindero.
\newblock Conditional generative adversarial nets.
\newblock {\em CoRR}, abs/1411.1784, 2014.

\bibitem{MooJanPetSch09}
Joris Mooij, Dominik Janzing, Jonas Peters, and Bernhard Sch\"{o}lkopf.
\newblock Regression by dependence minimization and its application to causal
  inference in additive noise models.
\newblock In {\em International Conference on Machine Learning (ICML)}, 2009.

\bibitem{QuaSha17}
Novi Quadrianto and Viktoriia Sharmanska.
\newblock Recycling privileged learning and distribution matching for fairness.
\newblock In {\em Advances in Neural Information Processing Systems (NIPS)},
  2017.

\bibitem{RahRec08}
Ali Rahimi and Benjamin Recht.
\newblock Random features for large-scale kernel machines.
\newblock In {\em Advances in Neural Information Processing Systems (NIPS)},
  2008.

\bibitem{SatHofChe18}
Prasanna Sattigeri, Samuel~C. Hoffman, Vijil Chenthamarakshan, and Kush~R.
  Varshney.
\newblock Fairness {GAN}.
\newblock {\em arXiv}, arXiv:1805.09910, 2018.

\bibitem{SimZis15}
K. Simonyan and A. Zisserman.
\newblock Very deep convolutional networks for large-scale image recognition.
\newblock In {\em International Conference on Learning Representations (ICLR)},
  2015.

\bibitem{SonSmoGreBedetal12}
Le Song, Alex Smola, Arthur Gretton, Justin Bedo, and Karsten Borgwardt.
\newblock Feature selection via dependence maximization.
\newblock {\em Journal of Machine Learning Research (JMLR)}, 13:1393--1434,
  2012.

\bibitem{UlyLebVedLem16}
Dmitry Ulyanov, Vadim Lebedev, Andrea Vedaldi, and Victor~S. Lempitsky.
\newblock Texture networks: Feed-forward synthesis of textures and stylized
  images.
\newblock In {\em International Conference on Machine Learning (ICML)}, 2016.

\bibitem{ulyanov2017}
Dmitry Ulyanov, Andrea Vedaldi, and Victor Lempitsky.
\newblock Improved texture networks: Maximizing quality and diversity in
  feed-forward stylization and texture synthesis.
\newblock In {\em Computer Vision and Pattern Recognition (CVPR)}, 2017.

\bibitem{WacMitRus18}
Sandra Wachter, Brent Mittelstadt, and Chris Russell.
\newblock Counterfactual explanations without opening the black box: Automated
  decisions and the gdpr.
\newblock {\em Harvard Journal of Law \& Technology}, 31(2), 2018.

\bibitem{RS2017}
{Working Group}.
\newblock Machine learning: the power and promise of computers that learn by
  example.
\newblock Technical report, The Royal Society, 2017.

\bibitem{ZafValRodGum17b}
Muhammad~Bilal Zafar, Isabel Valera, Manuel Gomez{-}Rodriguez, and Krishna~P.
  Gummadi.
\newblock Fairness beyond disparate treatment {\&} disparate impact: Learning
  classification without disparate mistreatment.
\newblock In {\em International Conference on World Wide Web (WWW)}, pages
  1171--1180, 2017.

\bibitem{ZemWuSwePitetal13}
Richard~S. Zemel, Yu Wu, Kevin Swersky, Toni Pitassi, and Cynthia Dwork.
\newblock Learning fair representations.
\newblock In {\em International Conference on Machine Learning (ICML)}, 2013.

\bibitem{ZhaLemMit18}
Brian~Hu Zhang, Blake Lemoine, and Margaret Mitchell.
\newblock Mitigating unwanted biases with adversarial learning.
\newblock In {\em Association for the Advancement of Artificial Intelligence
  (AAAI)}, 2018.

\bibitem{ZhuParIsoEfr17}
Jun-Yan Zhu, Taesung Park, Phillip Isola, and Alexei~A. Efros.
\newblock Unpaired image-to-image translation using cycle-consistent
  adversarial networks.
\newblock In {\em International Conference on Computer Vision (ICCV)}, 2017.

\end{thebibliography}
}

\end{document}


\title{Supplementary Material -- \\ Discovering Fair Representations in the Data Domain}


\maketitle

\section{Adult Dataset Experiments}

\subsection{Details}
We compare our method against an unmodified $\x{}$ using the following classifiers: 
1) logistic regression (\texttt{LR}) and 
2) support vector machine with linear kernel (\texttt{SVM}),
We select the regularization parameter of \texttt{LR} and \texttt{SVM} over 6 possible values using $3$-fold cross validation.
We then train classifiers 1--2 with the learned representation $\xtilde{}$ and with the latent embedding $\z{}$ of a state-of-the-art adversarial model described in Beutel et al. \cite{BeuCheZhaChi17}. 
As a reference, we also compare with:
3) Zafar et al.'s\cite{ZafValRodGum17b} method (\texttt{Zafar.}), a state-of-the-art method that adds equality of opportunity directly as a constraint to the classifier learning objective function.
We also apply methods which reweigh the samples to simulate a balanced dataset with regard to sensitive attributes
4) FairLearn \cite{Aga18} optimized with both the cross-validated \texttt{LR} 
and \texttt{SVM} (1\&2), (\texttt{FL LR}) and (\texttt{FL SVM}) respectively,
5) Kamiran \& Calders \cite{KamCal12} with both the cross-validated \texttt{LR} 
and \texttt{SVM} (1\&2) (\texttt{K\&C LR}) and (\texttt{K\&C SVM}) respectively

It has been shown that applying fairness constraints in succession as `fair pipelines' do not enforce fairness \cite{DwoIlv18, BowKitNisStraVarVen17}, as such, we only demonstrate (fair) classifier 3 on the unmodified $\x{}$.

\textbf{Benchmarking}
We follow the evaluation protocol of Friedler et al. \cite{FrieSchVen18}.
We evaluate the effect of styling the Adult dataset using a modified version of the framework provided by Friedler et al. in their work comparing fairness-enhancing interventions \cite{FrieSchVen18}. 
We train our model for $50,000$ iterations using a network with 1 hidden layer of $40$ nodes for both the encoder and decoder, with the encoded representation being 40 nodes. The predictor acts on the decoded output of this network. We set the trade-off parameters of the reconstruction loss ($\lambda_1$) and decomposition loss ($\lambda_2$) to $10^{-4}$ and $100$ respectively.
We then use this model to translate training and test sets into $\xtilde{}$ for $10$ repeated experiments, evaluating all the methods using $\x{}$ and $\xtilde{}$ representations. To ensure consistency, we train the model of Beutel et al. \cite{BeuCheZhaChi17} with the same architecture and number of iterations.
and number of randomized repeats.
on the training set to produce the training set for $\xtilde{}$ and use the styling model on the test set in $\x{}$ to produce the test set for $\xtilde{}$. We then train the models described above on the training set of both $\x{}$ and $\xtilde{}$ respectively, before evaluating the models on the corresponding test set.

\subsection{Results}

Experimental results over $10$ repeats are presented in Table \ref{table:benchmarking}.
The metric ``Eq. Opp.'' is measured as the absolute difference in true positive rates across the two protected groups.
Our learned representation, $\xtilde{}$ achieves a similar fairness level to Beutel's state-of-the-art approach (\textbf{Q1}). Consistently our translated representation $\xtilde{}$ promotes the \emph{fairness} criterion (Eq. Opp. close to $0$) for all classifiers, with only a small penalty in accuracy, while maintaining the constraint of being in the same space as the original input. 
Classifier 3, which is not constrained to learn a representation, outperforms both representation learning approaches, however the model only converged in 6 out of 10 repeats.

\begin{table*}[]
\centering
\scalebox{0.98}{%
\begin{tabular}{lccccccccc}
\toprule
{} & \multicolumn{2}{c}{original $\x$} & \multicolumn{2}{c}{fair interpretable $\xtilde$} & \multicolumn{2}{c}{latent embedding $\mathbf{z}$} \\
{}  & Accuracy $\uparrow$ &  Eq. Opp $\downarrow$  & Accuracy $\uparrow$ &  Eq. Opp $\downarrow$ & Accuracy $\uparrow$ &  Eq. Opp $\downarrow$ \\
\midrule
1: \texttt{LR}          &    $0.851\pm0.002$ &  $\mathbf{0.092\pm0.023}$  & $0.842\pm0.003$ &  $\mathbf{0.056\pm0.025}$  & $0.818\pm0.021$ &  $\mathbf{0.059\pm0.046}$ \\
2: \texttt{SVM}         &    $0.851\pm0.002$ &  $\mathbf{0.082\pm0.023}$  & $0.842\pm0.003$ &  $\mathbf{0.049\pm0.028}$  & $0.819\pm0.020$ &  $\mathbf{0.067\pm0.047}$ \\
3: \texttt{FL LR}  &    $0.851\pm0.002$ &  $\mathbf{0.149\pm0.013}$  & $0.841\pm0.003$ &  $\mathbf{0.065\pm0.032}$  & $0.818\pm0.021$ &  $\mathbf{0.056\pm0.048}$ \\
4: \texttt{FL SVM} &    $0.851\pm0.002$ &  $\mathbf{0.082\pm0.023}$  & $0.842\pm0.003$ &  $\mathbf{0.049\pm0.028}$  & $0.819\pm0.020$ &  $\mathbf{0.067\pm0.047}$ \\
5: \texttt{K\&C LR}     &    $0.844\pm0.002$ &  $\mathbf{0.149\pm0.013}$  & $0.841\pm0.003$ &  $\mathbf{0.017\pm0.013}$  & $0.818\pm0.021$ &  $\mathbf{0.049\pm0.033}$ \\
6: \texttt{K\&C SVM}    &    $0.851\pm0.002$ &  $\mathbf{0.082\pm0.023}$  & $0.842\pm0.003$ &  $\mathbf{0.049\pm0.028}$  & $0.819\pm0.020$ &  $\mathbf{0.067\pm0.047}$ \\
7: \texttt{Zafar.}*  & $0.850\pm0.003$ &  $\mathbf{0.018\pm0.009}$  &  --- &    ---  &      --- &    --- \\
\bottomrule
\end{tabular}
}

\caption{Results of training multiple classifiers (rows 1--7) including the fair classifier of Zafar et al. \cite{ZafValRodGum17b} on three different representations, $\x$, $\xtilde$, and $\mathbf{z}$. $\x$ is the original input representation, $\xtilde$ is the interpretable, fair representation introduced in this paper, and $\mathbf{z}$ is the latent embedding representation of Beutel et al. \cite{BeuCheZhaChi17}. We \textbf{boldface} Eq. Opp. since this is the fairness criterion (the lower the better). *\texttt{Zafar.} classifier failed to converge in 4 out of 10 repeats. Our learned representation $\xtilde$ achieves comparable fairness level to the latent representation $\mathbf{z}$, while maintaining the constraint of being in the same space as the original input.}
\label{table:benchmarking}
\end{table*}

{\small
\bibliographystyle{ieee_fullname}
\bibliography{bibfile}
}